\documentclass[letterpaper, 10 pt, conference]{ieeeconf}  

\usepackage{booktabs,multirow,makecell}             
\usepackage{graphicx}
\usepackage{multirow}
\usepackage{xcolor}
\usepackage{algorithm}
\usepackage{algpseudocode}
\usepackage{mathtools}
\usepackage{stfloats}
\usepackage{hyperref}  
\usepackage{subcaption}
\usepackage{amsmath}
\usepackage{multicol}
\usepackage{amssymb}
\usepackage[colorinlistoftodos,prependcaption,textsize=tiny]{todonotes}
\usepackage{cite}
\usepackage{capt-of}
\usepackage{caption}
\usepackage[none]{hyphenat}

\graphicspath{ {./} }

\ifdefined\providecolor\providecolor{bestINT8}{HTML}{FF7F0E}\providecolor{bestF32}{HTML}{0B7F87}\else\providecommand{\textcolor}[2]{#2}\fi

\makeatletter
\newcommand\fs@spaceruled{\def\@fs@cfont{\bfseries}\let\@fs@capt\floatc@ruled
  \def\@fs@pre{\vspace{0.5\baselineskip}\hrule height.8pt depth0pt \kern2pt}%
  \def\@fs@post{\kern2pt\hrule\vspace{-0.95\baselineskip}}
  \def\@fs@mid{\kern2pt\hrule\kern2pt}%
  \let\@fs@iftopcapt\iftrue}
\makeatother

\IEEEoverridecommandlockouts                              
\title{\LARGE \bf
TinyFlow: Breaking the Warping Bottleneck for Edge Optical Flow

\author{Xiao'ao Song and Chahat Deep Singh\\University of Colorado, Boulder
\thanks{This work has been submitted to the IEEE for possible publication.
Copyright may be transferred without notice, after which this version may
no longer be accessible.}%
\thanks{The authors are with the Perception, Robotics, AI, Sensing (PRAISe)
Lab, University of Colorado, Boulder, CO, USA.
Email: \{xiaoao.song, chahat.singh\}@colorado.edu.}}
}

\begin{document}
\IEEEaftertitletext{
    \begin{center}
        \includegraphics[width=1.0\textwidth]{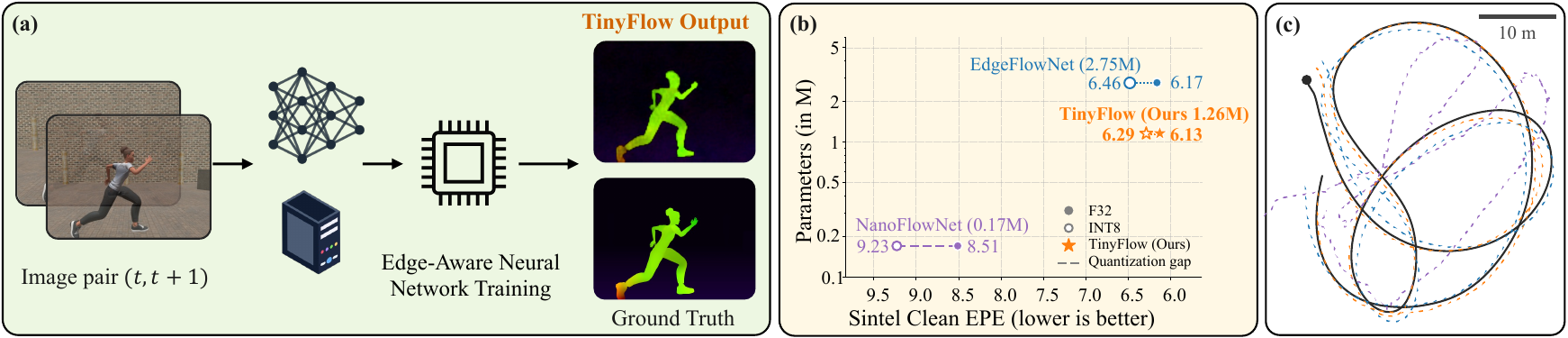}
        \captionof{figure}{%
        \textbf{TinyFlow.} 
        (a) Our warp-free optical flow network with only 1.26M parameters utilizes a pair of input images to predict dense, accurate optical flow on edge AI accelerators. 
        (b) Sintel Clean dataset: TinyFlow has the lowest EPE with 54\% fewer parameters than EdgeFlowNet and the smallest quantization gap 0.16 vs.\ 0.29 (EdgeFlowNet) / +0.72 (NanoFlowNet), evaluated on $384\times512$\,px
        images.
        (c) Flow-only monocular VO on real UZH-FPV dataset (colors as in (b), ground truth in black): TinyFlow INT8 reaches 2.04\,m ATE vs.\ 2.41\,m (EdgeFlowNet) and 15.84\,m (NanoFlowNet).
        }
        \label{fig:banner}
    \end{center}
}

\maketitle
\thispagestyle{empty}

\begin{abstract}
High-performance optical flow networks are difficult to deploy on resource-constrained edge accelerators because their iterative refinement warps features by the current flow, which relies on flow-dependent resampling and data-dependent memory access. We present TinyFlow, a 1.26M-parameter warp-free optical flow network that preserves iterative coarse-to-fine refinement by replacing the warp with a fixed-window correlation over constant offsets, so the whole network compiles as a single static INT8 graph. Quantization-aware fine-tuning, warp-teacher distillation, and model souping reduce the Float32-to-INT8 penalty to below 0.2 end-point error (EPE) on MPI-Sintel. TinyFlow outperforms EdgeFlowNet and NanoFlowNet on Sintel Clean, Sintel Final, and FlyingChairs in both Float32 and INT8, while using 54\% fewer parameters than EdgeFlowNet, and our INT8 model achieves 6.29/7.13 EPE on Sintel Clean/Final. In flow-only monocular visual odometry on TartanAir and UZH-FPV, the same INT8 model yields the lowest trajectory error on seven of eleven sequences.

\end{abstract}

\section{Introduction}
Before a robot can act on what it sees, it has to know what is moving, and dense optical flow answers that question at every pixel. Optical flow is crucial for navigation ~\cite{chao2014survey}, visual odometry ~\cite{pandey2021leveraging}, structure from
motion ~\cite{schonberger2016structure}, object tracking ~\cite{karpuzov2024object} and dynamic scene understanding ~\cite{luo2022motion} and much more. Learning-based estimators have raised its accuracy far beyond classical methods, but they are designed and evaluated on desktop GPUs. Their memory and compute cost place them out of reach of the onboard processors of small robots.

With the need for onboard perception growing, autonomous drones and mobile systems must perceive the world in real time on low-power hardware. Several works have pushed vision models such as optical flow and depth networks onto such platforms. Bouwmeester et al.~\cite{bouwmeester2023nanoflownet} ran dense optical flow on a Bitcraze Crazyflie nano quadcopter with a GAP8 microcontroller. Singh et al.~\cite{singh2024minimal} reached 4.5\,Hz depth prediction for palm-sized drones on a Raspberry Pi Compute Module~4. Raju et al.~\cite{edgeflownet2024} reported 100\,Hz optical
flow at \texttt{UINT8} precision on a Google Coral Edge TPU drawing 1.08\,W, and Krispin-Avraham et al.~\cite{krispin2024real} deployed 3D object detection on a Hailo-8 accelerator at roughly 5\,Hz.

Deployable optical flow has so far meant giving up the mechanism that makes optical flow accurate. Every network that runs on an edge accelerator today, like NanoFlowNet \cite{bouwmeester2023nanoflownet} and EdgeFlowNet \cite{edgeflownet2024}, is a single feed-forward pass, and every network at the top of the accuracy
benchmarks is iterative. Iterative estimators refine their prediction in a loop that warps features by the current flow before each correction, usually through \texttt{grid\_sample}, so that each step only has to
resolve the motion that remains. That warp reads memory at addresses computed from the data at run time, and an accelerator that executes a fixed dataflow graph has no way to perform it. Deployable networks therefore leave out
iterative refinement, and their lower accuracy follows from that omission rather than from their smaller size.

To address this limitation, we propose \textbf{TinyFlow}, a 1.26M-parameter optical flow network that keeps the iterative refinement loop and removes the warp. Instead of warping features by the current flow, TinyFlow correlates the two frames over a fixed window of 
constant offsets, so the flow estimate selects among channel values rather than steering memory addresses. The data dependence that makes warping undeployable does not disappear. It moves into ordinary arithmetic. As a result, the entire network exports, quantizes, and
compiles for edge AI accelerators (Fig.~\ref{fig:banner}).

The primary contributions are as follows:

\begin{enumerate}
\item We propose TinyFlow, a 1.26M-parameter warp-free optical flow network that retains iterative refinement and compiles to a single static INT8 graph.
\item We demonstrate that quantization-aware fine-tuning with warp-teacher distillation and model souping together reduce the Float32 to INT8 penalty from over 1.0 to below 0.2 EPE on Sintel dataset.
\item We show that TinyFlow outperforms EdgeFlowNet and NanoFlowNet on Sintel and FlyingChairs in both Float32 and INT8 with 54\% fewer parameters than EdgeFlowNet.
\end{enumerate}

\section{Related works}
\textbf{Iterative optical flow.} 
FlowNet~\cite{dosovitskiy2015flownet} established end-to-end flow learning. FlowNet 2.0~\cite{ilg2017flownet} stacked networks with warping between them and markedly improved accuracy. PWC-Net~\cite{sun2018pwc} made warping and cost volumes the standard coarse-to-fine toolkit. RAFT~\cite{teed2020raft} replaced the pyramid with a recurrent update operator that looks up a correlation volume at the current flow on every iteration. These designs share one property. Their accuracy comes from repeated alignment driven by the current flow estimate, and each implements that alignment as data-dependent resampling. TinyFlow keeps iterative refinement and removes the flow-dependent resampling.

\textbf{Lightweight and edge optical flow.} LiteFlowNet3~\cite{hui2020liteflownet3} and FastFlowNet~\cite{kong2021fastflownet} reduce the cost of the pyramid design but retain warping. DCVNet~\cite{jiang2023dcvnet} avoids warping
with dilated cost volumes that widen the search range at fixed cost. NeuFlow~\cite{zhang2025neuflow} targets larger embedded GPUs such as NVIDIA Jetson boards: NeuFlow v1 warps second-frame features with \texttt{grid\_sample}, and NeuFlow v2 replaces the warp with a flow-indexed \texttt{grid\_sample} lookup into
a correlation pyramid on every iteration. Both also build a global correlation with a dense matrix product and a softmax. These operations run well on a GPU-class device such as the Jetson Orin Nano that their paper targets, but none belongs to the vocabulary of a static-dataflow
INT8 accelerator. NanoFlowNet \cite{bouwmeester2023nanoflownet} runs on a GAP8 microcontroller at $112{\times}160$ and EdgeFlowNet~\cite{edgeflownet2024} reaches 100\,FPS at about one watt on a Coral Edge TPU and are the closest works to ours in terms of model size. Both fit an accelerator by removing warping and iteration entirely and predicting flow in a single pass. TinyFlow
instead re-expresses the refinement mechanism in a compile-legal form
and keeps the accuracy that comes with it.

\textbf{Quantization.} Jacob et~al.~\cite{jacob2018quantization} formalized integer-arithmetic inference with per-channel weight scales and fake quantization during training. Knowledge distillation~\cite{hinton2015distilling} transfers accuracy from a larger teacher, and model soups~\cite{wortsman2022model} average checkpoints in weight space. TinyFlow combines the three for deployment. Quantization-aware fine-tuning distills from a warp-based teacher that the accelerator itself cannot run, a soup averages the resulting checkpoints, and checkpoint selection uses quantized rather than float accuracy.
\section{TinyFlow Framework}
\label{sec:method}

\begin{figure*}[!t] 
    \centering
    \includegraphics[width=1.0\textwidth]{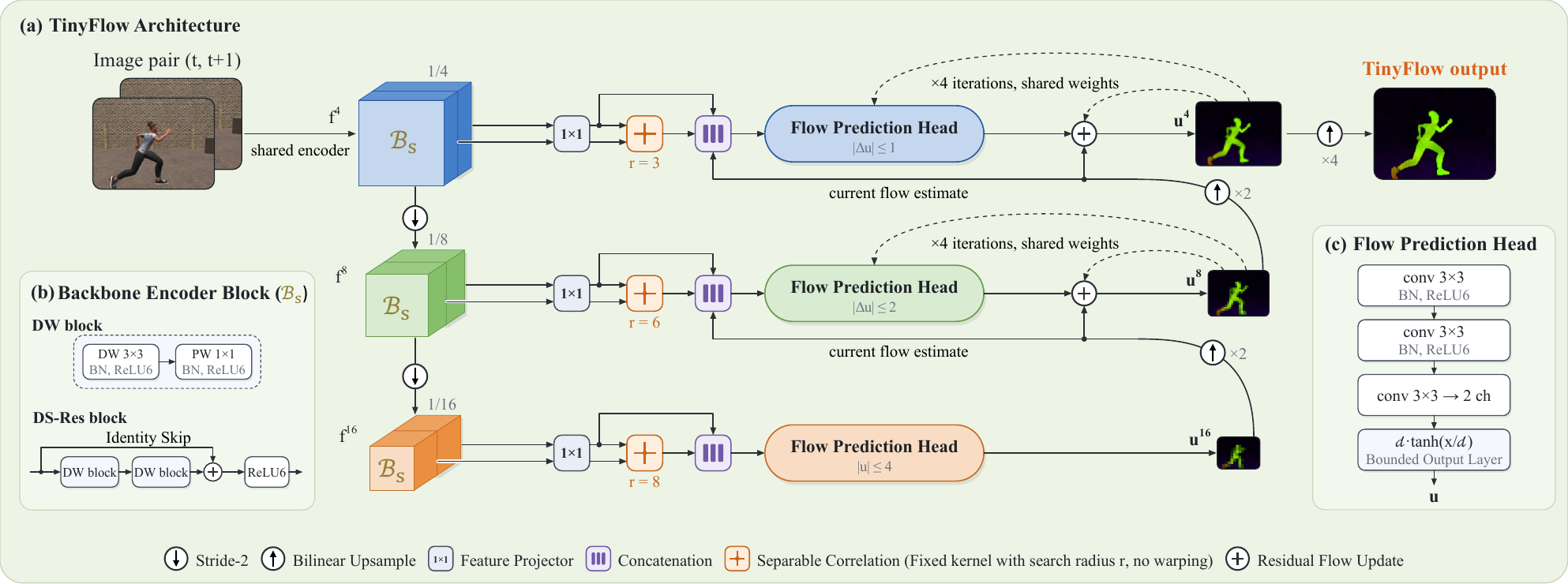}
    \caption{\textbf{TinyFlow architecture.} (a)~A shared encoder builds a multi-scale feature pyramid; separable correlation matches the features at each scale and the flow is refined coarse-to-fine. (b)~The backbone encoder is built from lightweight feature-extraction blocks of depth-wise and point-wise convolutions. (c)~The three flow prediction heads have the same architecture but separate weights.   }
    \label{fig:architecture}
\end{figure*}

\subsection{Problem Formulation}
A camera captures two consecutive RGB frames $I_1$ and $I_2$, each in $[0,1]^{H \times W \times 3}$. Let us say a pixel at position $\mathbf{x} = (x, y)$ in $I_1$ moves to position $\mathbf{x} + \mathbf{u}(\mathbf{x})$ in $I_2$, so
the optical flow $\mathbf{u}(\mathbf{x}) \in \mathbb{R}^2$ is its displacement
measured in pixels. We estimate the flow at three scales
$s \in \{16, 8, 4\}$, each downsampled by a factor of $s$ to $(H/s, W/s)$.
Estimation starts at the coarsest scale, and each finer scale refines the
result of the one before it.

\subsection{The Problem with Dynamic Resampling}
Various optical flow methods refine the flow by warping
features of the second frame under the current estimate. Let
$f^s_i \in \mathbb{R}^{C_s \times H_s \times W_s}$ denote the feature map that a learned encoder extracts from frame $I_i$ at scale $s$, with $C_s$ channels at the spatial resolution $H_s = H/s$ and $W_s = W/s$ of that scale. The warp is defined as
$\tilde{f}^s_2(\mathbf{x}) \triangleq f^s_2(\mathbf{x} +
\mathbf{u}^s(\mathbf{x}))$, which brings the second frame's features into alignment with the first before the next correction. Evaluating the right-hand side requires bilinear resampling at coordinates shifted by the flow, which are in general non-integer, and is commonly implemented with operators such as \texttt{grid\_sample}. 

Edge accelerators and Neural Processing Units or NPUs~\cite{reuther2022survey} handle this operation poorly. It requires memory reads that depend on the input, together with spatial sampling primitives that compilation toolchains frequently do not support because the access pattern is not fixed at compile time~\cite{bouwmeester2023nanoflownet, edgeflownet2024}. When the operator is not supported, the network either fails to compile or the warp executes on the host CPU and the accelerator's throughput advantage is lost. \emph{Our goal is therefore to retain coarse-to-fine flow estimation while building an inference graph that compiles for edge accelerators, without any need for dynamic resampling in the forward pass.}

Fig.~\ref{fig:architecture} shows the architecture that meets this goal.

\subsection{Multi-Scale Feature Representation}

A shared encoder $\mathcal{B}_s$ produces the feature pyramid introduced above, $f_i^s = \mathcal{B}_s(I_i)$, from each input image $I_i$, with $i \in \{1, 2\}$ indexing the frame and $s \in \{16, 8, 4\}$ the scale.
Its layers are depthwise-separable convolutions with pointwise
nonlinearities, which keep the parameter count small and quantize well.

Correlation is sensitive to the magnitude of the features it compares. The usual remedy is to normalize each pixel's feature vector, $
\widehat{f}_i^s(\mathbf{x}) =
f_i^s(\mathbf{x}) / (\lVert f_i^s(\mathbf{x}) \rVert_2 + \varepsilon).$

Although this normalization is effective in floating-point training, it introduces channel reduction plus nonlinear and data-dependent normalization that are less hardware-friendly. Therefore, before computing correlation, we pass the features through a small learnable projector,
\begin{equation}
\hat{\mathbf{f}} = \cfrac{\mathrm{BN}(\mathrm{Conv}_{1\times1}\mathbf{f})}{\sqrt{C}}
\end{equation}
where $C$ is the number of channels at that scale. The projector serves the same purpose and holds the correlation values inside a bounded range. Batch normalization limits the magnitude of each feature, and the division by $\sqrt{C}$ removes the growth of an inner product with the number of channels. The $1{\times}1$ convolution starts as an identity, so training begins from plain correlation and departs from it only when a learned projection helps. This is also quantization-friendly, since the convolution, the batch normalization, and the fixed scale fuse into one integer-legal operation. All correlations in this paper are computed after this projection unless stated otherwise.

\subsection{Warp-Free Matching via Fixed-Window Correlation}
The next step is to match the projected pyramids across the two frames
without any flow-dependent resampling. Rather than building a residual cost
volume~\cite{sun2018pwc} around warped features, we correlate the two feature
maps directly over a fixed window of offsets,
\begin{equation}
C^s_0(\mathbf{x}, \boldsymbol{\delta}) \triangleq
\left\langle \hat{f}^s_1(\mathbf{x}),\,
\hat{f}^s_2(\mathbf{x} + \boldsymbol{\delta}) \right\rangle,
\qquad \boldsymbol{\delta} \in \mathcal{D}(r_s),
\label{eq:fixed_corr}
\end{equation}
where $\langle \cdot,\cdot \rangle$ is the inner product over channels and
$\mathcal{D}(r_s) = \{-r_s,\ldots,r_s\}^2$ is the offset set. In practice we
compute only the two axis marginals of this volume,
\begin{equation}
C^s_x(\mathbf{x}, \delta) = C^s_0(\mathbf{x}, (\delta, 0)), \quad
C^s_y(\mathbf{x}, \delta) = C^s_0(\mathbf{x}, (0, \delta))
\label{eq:sep_corr}
\end{equation}
where $\delta \in \{-r_s,\ldots,r_s\}$. So every spatial location carries $K_s = 2(2r_s{+}1)$ correlation channels rather than $(2r_s{+}1)^2$, and the update block learns to combine the two
marginals. Every offset is a constant fixed at compile time, so the volume is an array of plainly shifted inner products and involves no data-dependent addressing.

\subsubsection{Relationship to Warping}
A warped residual cost volume has the form
\begin{equation}
C^s_{\mathbf{u}}(\mathbf{x}, \boldsymbol{\delta}) \triangleq
\left\langle \hat{f}^s_1(\mathbf{x}),\,
\hat{f}^s_2(\mathbf{x} + \mathbf{u}^s(\mathbf{x}) + \boldsymbol{\delta}) \right\rangle.
\label{eq:warped_corr}
\end{equation}
For integer flow, $\mathbf{u}^s(\mathbf{x}) \in \mathbb{Z}^2$, this volume is
a shifted read of the fixed one,
\begin{equation}
C^s_{\mathbf{u}}(\mathbf{x}, \boldsymbol{\delta}) =
C^s_0(\mathbf{x}, \boldsymbol{\delta} + \mathbf{u}^s(\mathbf{x})).
\label{eq:shift_identity}
\end{equation}
For the full window, warping therefore adds no matching evidence that the
fixed volume does not already contain. Its only effect is to recenter the
search window on the current estimate. For the axis marginals we compute,
the same holds along each axis when the orthogonal component of the residual
displacement is small relative to the smoothness of the features across that
axis, which the coarse-to-fine schedule below is designed to keep true. We
keep the window centered at zero and make the radii $\{r_s\}$ large enough
that the true displacement does not leave it. The design question is how
large the radii must be.

\subsubsection{Capture Range}
One correlation offset at scale $s$ corresponds to $s$ pixels at full
resolution, so a radius $r_s$ reaches $\pm r_s s$ pixels along each axis.
The window contains the true displacement $\mathbf{u}^{\mathrm{full}}(\mathbf{x})$
whenever both of its components are at most $r_s s$ in magnitude, that is
$\|\mathbf{u}^{\mathrm{full}}(\mathbf{x})\|_\infty \le r_s s$.
Because only the axis marginals are computed, the match is found within
that window only if the features also vary slowly across the orthogonal
axis over the residual displacement. Coarse-to-fine refinement provides
both conditions. Each scale starts from the upsampled coarser estimate, so
its window only has to capture the residual displacement that estimate
leaves behind, and that residual is smaller at every level. Large radii
therefore belong to the coarse scales, where the grid is small and the added
correlation channels are cheap, while the fine scales use small radii. We
use $r_{16}{=}8$, $r_8{=}6$ and $r_4{=}3$, which cover $\pm128$, $\pm48$ and
$\pm12$ pixels along each axis respectively.

\subsection{Coarse-to-Fine Warp-Free Flow Estimation}
The cost volumes of the previous section supply the matching evidence at different scales $s$. This section combines these matches into optical flow. Estimation begins at the coarsest scale. The cost volume of Eq. \eqref{eq:sep_corr} is concatenated with the first-frame features, $\mathbf{x}^{16} = [C^{16}_0, f^{16}_1]$, and a small convolutional head $h_{16}$ regresses the initial flow $\mathbf{u}^{16} = d_{16} \cdot \tanh\big(h_{16}(\mathbf{x}^{16}) /
d_{16}\big)$. The scaled hyperbolic tangent limits the output to
$\pm d_{16}$ pixels.

Each finer scale $s \in \{8, 4\}$ starts from the estimate above it, upsampled to initialize $\mathbf{u}^s_0 = 2 \cdot \mathrm{Up}(\mathbf{u}^{2s})$, where $\mathrm{Up}(\cdot)$ is bilinear upsampling and the factor of two keeps the flow in pixel units of the finer grid. A weight-tied update block $g_s$ then refines the estimate over $t = 0, \ldots, T_s - 1$ steps, with $T_s = 4$ during training. Every step regresses a bounded correction and adds it to the running estimate, 
$\mathbf{u}^s_{t+1} = \mathbf{u}^s_t + \Delta\mathbf{u}^s_t$, with: 
\begin{equation}
\Delta\mathbf{u}^s_t = d_s \cdot
\tanh\big(g_s([C^s_0,\, f^s_1,\, \mathbf{u}^s_t]) / d_s\big)
\label{eq:update}
\end{equation}

The cost volume $C^s_0$ is computed once per scale and is never
recentered. The current flow enters the update block as an input channel
instead, so the data dependence that warping places in memory addresses is
carried in feature values, which is ordinary arithmetic. Each correction
is limited to $\pm d_s$ pixels in the same way as the initial flow, with
$d_{16} = 4$, $d_8 = 2$ and $d_4 = 1$. Because every step is bounded, the
total flow a scale can add is at most $T_s d_s$ pixels, so the numeric
range of every flow tensor is known before training. This keeps the
regression stable and makes the tensors representable in 8-bit encodings.
The final prediction is $\mathbf{u}^4_{T_4}$ at quarter resolution, which
is bilinearly upsampled to full resolution for evaluation.


\subsection{Edge-Compilable Correlation}
The estimator rests on the marginals of Eq.~\eqref{eq:sep_corr}, and it is
deployable because they need no dynamic indexing. Gathering a
$(2r_s{+}1)$-wide strip of $\hat{f}^s_2$ around every pixel exports as an
indexing operator that edge toolchains reject. A shift by a constant offset
is instead a convolution with a one-hot kernel, so each marginal becomes
three fixed-weight layers. A convolution with one-hot kernels produces the
$2r_s{+}1$ shifted copies of every channel of $\hat{f}^s_2$, laid out
offset-major so that copy $kC_s{+}c$ holds channel $c$ at offset $k$. A
$1{\times}1$ convolution with one-hot weights replicates $\hat{f}^s_1$ in
the same layout, an elementwise product multiplies the two, and a grouped
$1{\times}1$ convolution with unit weights and $2r_s{+}1$ groups sums the
$C_s$ channels of each offset, which is exactly Eq.~\eqref{eq:sep_corr}.
The layout is essential. A depthwise shift convolution emits the copies
channel-major, with the offset varying fastest, and the same grouped
reduction then sums contiguous blocks that span every offset of a few
channels, so each output is a product of $\hat{f}^s_1$ with a box-filtered
$\hat{f}^s_2$ rather than an inner product at one offset. No weight in the
construction is learned and no memory address depends on the input. To the
compiler the volume is ordinary \texttt{Conv} and \texttt{Mul} layers; the
one-hot shift compiles to a dense layer of $C_s^2(2r_s{+}1)^2$ weights,
which is the cost that keeps large windows at the coarse scales.


\subsection{Training Objective}
\begingroup
\setlength{\abovedisplayskip}{3pt}\setlength{\belowdisplayskip}{3pt}
Let $\mathbf{u}^\star$ be the ground-truth flow at full resolution. Its version at scale $s$ is $\mathbf{u}^{\star,s} = \frac{1}{s}\mathrm{Down}_s(\mathbf{u}^\star)$, where $\mathrm{Down}_s$ is bilinear downsampling to the $H_s{\times}W_s$ grid and the division converts the values to that grid's pixel units. $\mathbf{u}^s$ denotes the final estimate at scale $s$, which is $\mathbf{u}^s_{T_s}$ for $s \in \{8, 4\}$.

\subsubsection{Robust Supervision} Each flow component is penalized with the Charbonnier function $\rho(z) = (z^2 + \epsilon^2)^{\gamma}$, with $\epsilon = 10^{-3}$ and $\gamma = 0.45$, which behaves like $L_1$ on large errors and is smooth at zero. Summed over the two components and averaged over the pixel grid $\Omega_s$, the data loss at scale $s$ is
\begin{equation}
\mathcal{L}_s = \frac{1}{|\Omega_s|} \sum_{\mathbf{x} \in \Omega_s}
\sum_{c \in \{u, v\}} \rho\big(u^s_c(\mathbf{x}) - u^{\star,s}_c(\mathbf{x})\big).
\end{equation}

\subsubsection{Deep Supervision} Every iterate $\mathbf{u}^s_t$ of Eq.~\eqref{eq:update} is supervised, so each step of the weight-tied block receives its own gradient, with later iterates weighted more,
\begin{equation}
\mathcal{L}_{\mathrm{ds},s} = \frac{1}{T_s} \sum_{t=1}^{T_s}
\frac{t}{T_s}\, \mathcal{L}_s(\mathbf{u}^s_t).
\end{equation}

\subsubsection{Edge-Aware Smoothness}
Let $G_4$ be the grayscale first frame bilinearly downsampled to quarter resolution, $\partial_d$ the finite-difference gradient along $d\in\{x,y\}$, and $w_d(\mathbf{x})=e^{-\beta|\partial_d G_4(\mathbf{x})|}$ an edge weight with $\beta=10$. The finest flow is regularized by
\begin{equation}
\mathcal{L}_{\mathrm{smooth}} = \frac{1}{2|\Omega_4|}\sum_{\mathbf{x}\in\Omega_4}\sum_{d\in\{x,y\}} w_d(\mathbf{x})\,\|\partial_d \mathbf{u}^{4}(\mathbf{x})\|_1,
\label{eq:smooth}
\end{equation}

which favors smooth flow in homogeneous regions and relaxes the penalty across image edges.

\noindent The total loss is therefore: 
\begin{equation}
\mathcal{L} = \sum_{s} \lambda_s \mathcal{L}_s +
\mu_8\mathcal{L}_{\mathrm{ds},8} + \mu_4\mathcal{L}_{\mathrm{ds},4} +
\nu\mathcal{L}_{\mathrm{smooth}},
\label{eq:total}
\end{equation}

with $\lambda_s = 1$ for all scales, $\mu_8 = 0.05$, $\mu_4 = 0.1$ and $\nu = 0.1$. The fine-tuning stage described later adds a distillation term. 
\endgroup

\subsection{Quantization-Aware Training (QAT)}
\begingroup
\setlength{\abovedisplayskip}{3pt}\setlength{\belowdisplayskip}{3pt}
The objective of Eq. \eqref{eq:total} yields an accurate Float32 network, but the deployed network runs in INT8. Quantizing a trained checkpoint after the fact is not sufficient. 
Post-training quantization of our base checkpoint raises Sintel Clean EPE (in px) from 6.92 in Float32 to 8.09 in INT8, a penalty of 1.17, because weights trained in floating point have never had to tolerate rounding.
We therefore fine-tune under simulated quantization. Fake-quantization nodes are inserted at exactly the points where the accelerator compiler will quantize, with per-channel symmetric INT8 weights and per-tensor asymmetric UINT8 activations, and gradients pass through them with the straight-through estimator~\cite{jacob2018quantization}. The weights adapt to 8-bit arithmetic during training rather than meeting it for the first time at export.

Two further measures act on the same stage: (i) First, the quantized network learns from a teacher as well as from ground truth. The teacher is a warped variant of TinyFlow that performs \texttt{grid\_sample} warping. It is more accurate than the warp-free network, and although it cannot be deployed, it can guide training. Following knowledge distillation \cite{hinton2015distilling}, we penalize the distance between
the two quarter-resolution flows,
\begin{equation}
\mathcal{L}_{\mathrm{dist}} = \frac{1}{2|\Omega_4|}
\sum_{\mathbf{x} \in \Omega_4}
\lVert \mathbf{u}^4(\mathbf{x}) - \mathbf{u}^4_{\mathrm{G}}(\mathbf{x}) \rVert_1,
\end{equation}
where $\mathbf{u}^4_{\mathrm{G}}$ is the teacher's flow. This term is added to Eq. \eqref{eq:total} with weight $\lambda_{\mathrm{dist}}$ between 0.25 and 0.5 across the fine-tuning stages. (ii) Second, float and
quantized accuracy do not rank checkpoints the same way, so we select checkpoints and average them \cite{wortsman2022model} by their INT8 accuracy rather than their float accuracy. Training uses FlyingChairs throughout, with FlyingThings3D added for fine-tuning. MPI-Sintel is used only for testing, never for training, fine-tuning, or calibration. Together, these measures reduce the quantization penalty from over $+1.0$ EPE to below $+0.2$ EPE. Sintel Clean EPE increases only from 6.13 to 6.29 and Final from 6.95 to 7.13. 
\endgroup

\section{Results}

\subsection{Training Setup}
We train our TinyFlow model with a learning rate of \mbox{$2\times10^{-4}$} and run quantization-aware fine-tuning at \mbox{$6\times10^{-5}$}. All experiments use a desktop PC with an AMD Ryzen~7 9800X3D CPU, 32~GB of RAM, and an NVIDIA GeForce\texttrademark{} RTX~5080 GPU. We report end point error in px on Sintel Clean, Sintel Final, and the FlyingChairs validation split, together with parameter counts. Finally, we test the models on a Hailo-8 AI accelerator mounted via PCIe on a Raspberry Pi 5 with 16 GB of RAM. Quantization uses the standard Hailo optimization (optimization level 1) with Hailo Dataflow Compiler (DFC) 3.34.0, and the quantized models run on HailoRT 4.23.0.


\subsection{Training Mixture, Model Souping, and Inference} Three further steps improve accuracy without changing the architecture. (i) First, FlyingThings3D~\cite{mayer2016things3d} is mixed into the  FlyingChairs \cite{dosovitskiy2015flownet} fine-tuning data. Its larger displacements supply supervision that FlyingChairs lacks, and at deploy iterations this stage lowers Sintel Clean EPE from 6.58 to 6.13. (ii) Second, we build a model soup~\cite{wortsman2022model} by averaging the weights of the three best fine-tuning epochs into a single checkpoint. The soup outperforms each of its ingredients on both passes. The three epochs score 6.27, 6.37 and 6.21 on Sintel Clean and 7.21, 7.06 and 7.12 on Final, while their average scores 6.13 and 6.95. (iii) Third, we increase the number of refinement iterations at inference. The update blocks at the 1/8 and 1/4 scales are weight-tied, so the same weights are applied more times with no retraining. Training uses four iterations at each scale and we deploy with eight, which lowers Sintel Clean EPE by about 0.5 while leaving the final pass unchanged. Beyond eight the Clean pass gains only marginally and the Final pass degrades, so eight is the operating point.

\subsection{Optical Flow Benchmark Evaluation}
\begin{table}[t!]
\centering
\caption{Trained on FlyingChairs and FlyingThings3D. Tested on MPI-Sintel and FlyingChairs validation set. Reference models are shown for context and are outside the deployable comparison. Underlined values mark the best reference-model result in each column. Bold marks the best deployable value in each column within each precision, teal for F32 and orange for INT8.}
\label{tab:flow_accuracy}

\renewcommand{\arraystretch}{1.08}
\setlength{\tabcolsep}{3.5pt}

\resizebox{\columnwidth}{!}{%
\begin{tabular}{l c c c c c c}
\toprule
& & &
\multicolumn{3}{c}{EPE$\downarrow$}
& \\
\cmidrule(lr){4-6}

Model
& Params
& Prec.
& \shortstack{Sintel\\Clean}
& \shortstack{Sintel\\Final}
& Chairs
& \shortstack{$>3$px\\(\%)} \\
\midrule

\multicolumn{7}{l}{
\textit{Large reference models. Shown for context.}
} \\
FlowNetS
& 38.7M
& F32
& 4.86
& 6.38
& 1.85
& 22.9
\\
NeuFlow v2
& 9.03M
& F32
& \underline{1.59}
& \underline{2.91}
& \underline{1.27}
& \underline{6.4}
\\
\midrule
\multicolumn{7}{l}{
\textit{Deployable models. Compile to a single static INT8 graph.}
} \\
\multirow{2}{*}{EdgeFlowNet}
& \multirow{2}{*}{2.75M}
& F32
& 6.17
& 7.30
& 3.59
& 21.2
\\
&
&
INT8
& 6.46
& 7.62
& 3.91
& 22.1
\\[2pt]

\multirow{2}{*}{NanoFlowNet$^\dagger$}
& \multirow{2}{*}{0.17M}
& F32
& 7.71
& 8.76
& 6.02
& 38.8
\\
&
&
INT8
& 7.99
& 8.68
& 6.28
& 40.7
\\[2pt]


\multirow{2}{*}{\textbf{TinyFlow (Ours)}} 
& \multirow{2}{*}{\textbf{1.26M}} 
& F32 
& \textbf{\textcolor{bestF32}{6.13}} 
& \textbf{\textcolor{bestF32}{6.95}} 
& \textbf{\textcolor{bestF32}{2.66}} 
& \textbf{\textcolor{bestF32}{20.6}} \\
& & INT8 
& \textbf{\textcolor{bestINT8}{6.29}} 
& \textbf{\textcolor{bestINT8}{7.13}} 
& \textbf{\textcolor{bestINT8}{2.71}} 
& \textbf{\textcolor{bestINT8}{21.1}} \\

\bottomrule
\end{tabular}%
}
{\footnotesize $^{\dagger}$All models are evaluated on $384\times512$\,px
images except NanoFlowNet, which is evaluated at its native resolution of
$112\times160$\,px. At $384{\times}512$ NanoFlowNet collapses to $20.7/22.5$ EPE respectively.\\
}
\end{table}

Table~\ref{tab:flow_accuracy} shows that TinyFlow outperforms EdgeFlowNet on every benchmark in both precisions while using 54\% fewer parameters (1.26M against 2.75M). Under INT8 it reaches 6.29/7.13 EPE on Sintel Clean and Final against 6.46/7.62 for EdgeFlowNet, and its FlyingChairs EPE is 31\% lower (2.71 against 3.91). TinyFlow also leads NanoFlowNet by a wide margin, even though NanoFlowNet is evaluated at its most favorable setting, the native $112{\times}160$ input, because its accuracy collapses at $384{\times}512$. FlowNetS is more accurate on the mean but uses about $30\times$ more parameters (38.7M), and NeuFlow v2 is more accurate still but depends on flow-dependent resampling, so neither can be deployed on the edge accelerators that this paper targets. Among the deployable models, TinyFlow offers the best trade-off between accuracy and size.
The $>3$\,px metric reports the percentage of pixels whose EPE exceeds 3\,px, with lower values indicating fewer large-error predictions. TinyFlow achieves 20.6\% and 21.1\% under F32 and INT8, respectively. These are the second-best results overall, behind only NeuFlow v2.

\begin{table}[t!]
\centering
\caption{EPE by ground-truth motion magnitude (px) on Sintel Clean and Final, fine bins. Every model except NanoFlowNet receives a $384{\times}512$ input and EPE is measured at the full ground-truth resolution, pixel-weighted within each bin. NanoFlowNet is evaluated at its native resolution of  $112{\times}160$px. Reference models are shown for context and are outside the deployable comparison. Underlined values mark the best reference-model result in each column. Bold marks the best deployable value in each column within each precision, teal for F32 and orange for INT8.}
\label{tab:bins}
\small
\setlength{\tabcolsep}{3.8pt}
\begin{tabular}{@{}l l r r r r r@{}}
\toprule
Model & Prec. & $s_{0\text{-}10}$ & $s_{10\text{-}40}$ & $s_{40\text{-}80}$ & $s_{80\text{-}128}$ & $s_{128+}$ \\
\midrule
\multicolumn{7}{c}{\textbf{Sintel Clean}} \\
\midrule
FlowNetS & F32 & 1.30 & 6.35 & 16.45 & 30.67 & 98.69 \\
NeuFlow v2 & F32 & \underline{0.45} & \underline{2.01} & \underline{4.79} & \underline{8.67} & \underline{40.97} \\
\midrule
\multirow{2}{*}{EdgeFlowNet} & F32 & \textbf{\textcolor{bestF32}{1.11}} & 7.10 & \textbf{\textcolor{bestF32}{21.42}} & 53.56 & 146.6 \\
 & INT8 & 1.39 & 7.43 & \textbf{\textcolor{bestINT8}{21.67}} & 53.88 & 147.2 \\
\multirow{2}{*}{NanoFlowNet} & F32 & 1.20 & 9.82 & 40.11 & 71.18 & 154.2 \\
 & INT8 & 1.54 & 11.25 & 42.19 & 72.97 & 154.0 \\
\multirow{2}{*}{\textbf{TinyFlow}} & F32 & 1.24 & \textbf{\textcolor{bestF32}{6.06}} & 21.86 & \textbf{\textcolor{bestF32}{51.99}} & \textbf{\textcolor{bestF32}{137.2}} \\
 & INT8 & \textbf{\textcolor{bestINT8}{1.32}} & \textbf{\textcolor{bestINT8}{6.20}} & 22.18 & \textbf{\textcolor{bestINT8}{53.30}} & \textbf{\textcolor{bestINT8}{137.6}} \\
\midrule
\multicolumn{7}{c}{\textbf{Sintel Final}} \\
\midrule
FlowNetS & F32 & 1.78 & 9.27 & 21.77 & 35.37 & 111.2 \\
NeuFlow v2 & F32 & \underline{0.64} & \underline{3.41} & \underline{10.56} & \underline{16.88} & \underline{78.71} \\
\midrule
\multirow{2}{*}{EdgeFlowNet} & F32 & \textbf{\textcolor{bestF32}{1.47}} & 9.23 & 26.17 & 58.15 & 147.2 \\
 & INT8 & \textbf{\textcolor{bestINT8}{1.73}} & 9.71 & 26.27 & 59.12 & 147.9 \\
\multirow{2}{*}{NanoFlowNet} & F32 & 1.74 & 11.31 & 38.30 & 72.42 & 155.5 \\
 & INT8 & 2.04 & 12.55 & 42.19 & 75.07 & 154.3 \\
\multirow{2}{*}{\textbf{TinyFlow}} & F32 & 1.70 & \textbf{\textcolor{bestF32}{8.95}} & \textbf{\textcolor{bestF32}{25.19}} & \textbf{\textcolor{bestF32}{51.78}} & \textbf{\textcolor{bestF32}{132.8}} \\
 & INT8 & 1.80 & \textbf{\textcolor{bestINT8}{8.91}} & \textbf{\textcolor{bestINT8}{25.23}} & \textbf{\textcolor{bestINT8}{54.94}} & \textbf{\textcolor{bestINT8}{133.5}} \\
\bottomrule
\end{tabular}
\end{table}

Table~\ref{tab:bins} separates the aggregate by ground-truth motion magnitude. 
The first two bins ($s_{0\text{-}10}$ and $s_{10\text{-}40}$) hold roughly 90\% of Sintel pixels. On Sintel Clean, under INT8 TinyFlow leads every deployable model on both, beating EdgeFlowNet by 0.07 EPE below 10 px and by 1.23 EPE between 10 and 40 px, where it also edges the $30\times$ larger FlowNetS. NanoFlowNet is the worst across all bins, but degrades the least beyond 128 px, suggesting that its error is spread more evenly rather than growing sharply with motion magnitude.
EdgeFlowNet retains a small edge in the 40 to 80 px bin, by 0.51 EPE, and TinyFlow leads again from 80 to 128 px and beyond. On Sintel Final, the advantage of TinyFlow at large motion is even more pronounced. Except for the $<10$ px bin, TinyFlow outperforms the other deployable models across every motion range in both F32 and INT8, with the gap generally widening as motion increases. In the 80--128 px bin, TinyFlow reduces EPE relative to EdgeFlowNet by 6.37 in F32 and 4.18 in INT8. Beyond 128 px, the margin over EdgeFlowNet reaches 14.4 EPE in both precisions. It also beats NanoFlowNet by retaining substantial margins of 22.7 EPE in F32 and 20.8 EPE in INT8. On both passes, FlowNetS keeps its advantage only above 40 px, and NeuFlow v2 leads every bin by a wide margin, which locates the cost of the deployment  constraint in the correlation window's reach rather than in capacity.

\begin{figure*}[t!]
\centering
\includegraphics[width=1.0\textwidth]{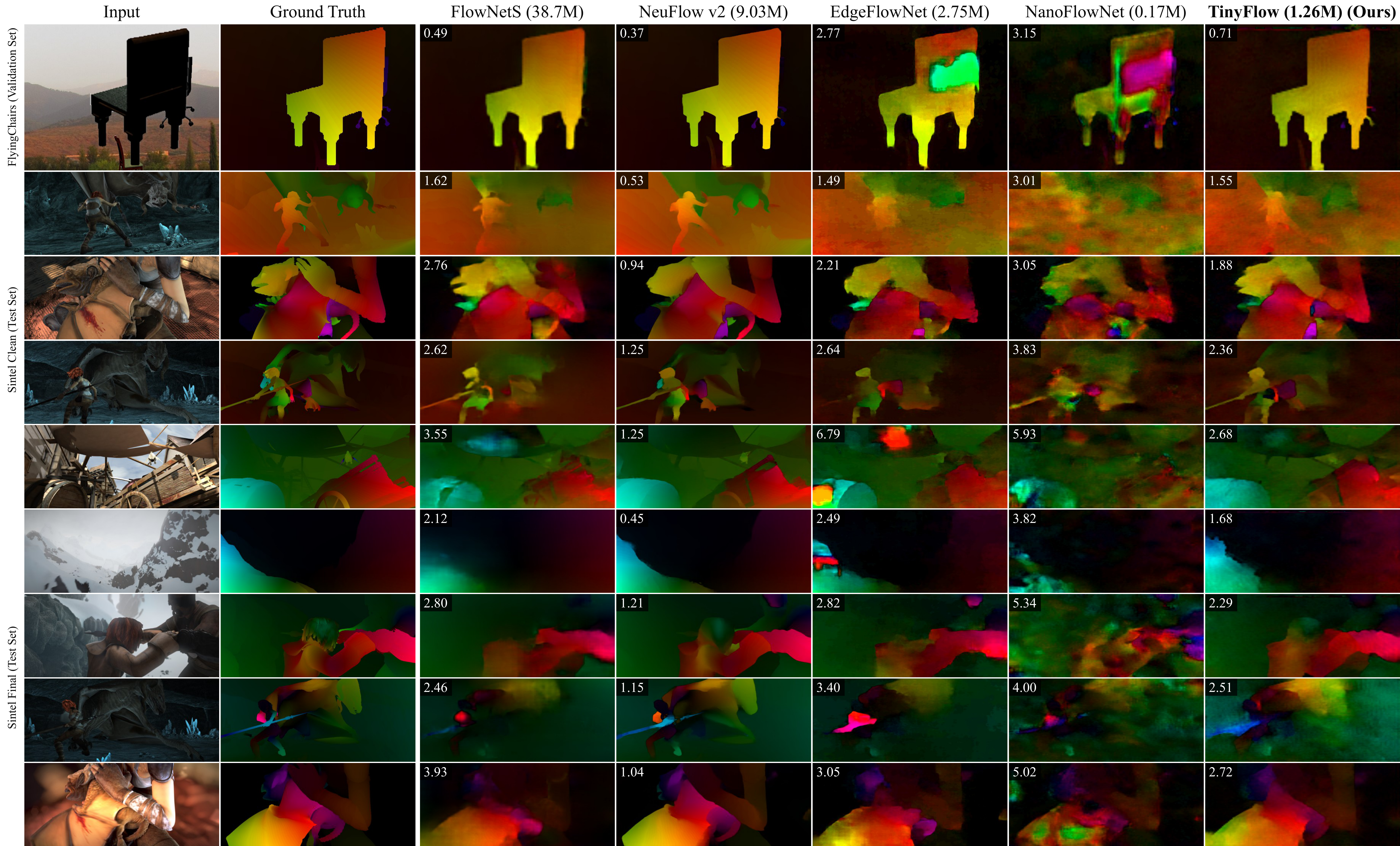}
\caption{\textbf{Qualitative comparison.} Hue encodes flow direction and brightness encodes magnitude, normalized per row to the ground truth's scale. The first column shows the inputs, with the first row from FlyingChairs validation, followed by four rows each from the Sintel Clean and Final test sets, and the second column shows the corresponding ground-truth flow. The remaining columns show FlowNetS, NeuFlow v2, EdgeFlowNet, NanoFlowNet, and TinyFlow (ours), each rendered by the predictor that produced its numbers in Table~\ref{tab:flow_accuracy}. FlowNetS and NeuFlow v2 are reference models outside the deployable  comparison. Rows are the lowest-EPE frame of each scene under TinyFlow, chosen among scenes with visible object motion. TinyFlow recovers the cleanest and sharpest motion segments among the deployable models at the smallest parameter count, EdgeFlowNet fragments large motion, and NanoFlowNet produces structured noise.}
\label{fig:fig_qual}
\end{figure*}

Fig.~\ref{fig:fig_qual} qualitatively illustrates the differences in the flow fields. On FlyingChairs, TinyFlow preserves the overall object silhouette and produces a spatially coherent flow field close to the ground truth. EdgeFlowNet introduces a large erroneous motion region on the upper part of the object, while NanoFlowNet produces strongly inconsistent and noisy flow within the object. 
Across the four Sintel Clean rows, TinyFlow preserves coherent motion regions with sharp boundaries. On \texttt{bandage\_1} (second Clean row), it keeps the overlapping bodies as separate motion regions, while NanoFlowNet produces a blurred and noisy flow field. The advantage is most pronounced on \texttt{market\_6} (fourth Clean row), where TinyFlow preserves the large foreground motion that EdgeFlowNet fragments into inconsistent patches, reaching 2.68 EPE against 6.79. NanoFlowNet also shows substantial structured noise across the Clean examples and has higher EPE than TinyFlow on all four rows. Despite having about $30\times$ fewer parameters, TinyFlow also achieves lower EPE than FlowNetS on all four Sintel Clean rows.
The four Sintel Final rows add motion blur and atmospheric effects. TinyFlow keeps clear and coherent motion boundaries on \texttt{ambush\_5} (second Final row), while NanoFlowNet produces blurred fields with strong structured noise. On \texttt{bandage\_1} (fourth Final row), TinyFlow preserves the foreground motion more coherently, whereas EdgeFlowNet introduces visible fragmentation and NanoFlowNet remains noisy. Across all four Final rows, TinyFlow achieves lower EPE than both EdgeFlowNet and NanoFlowNet. It also reaches lower EPE than FlowNetS on three of the four rows despite having about $30\times$ fewer parameters. NeuFlow v2 comes closest to the ground truth on every row, illustrating the accuracy gained when deployment constraints are removed.

\begin{figure*}[t!]
\centering
\includegraphics[width=1.0\textwidth]{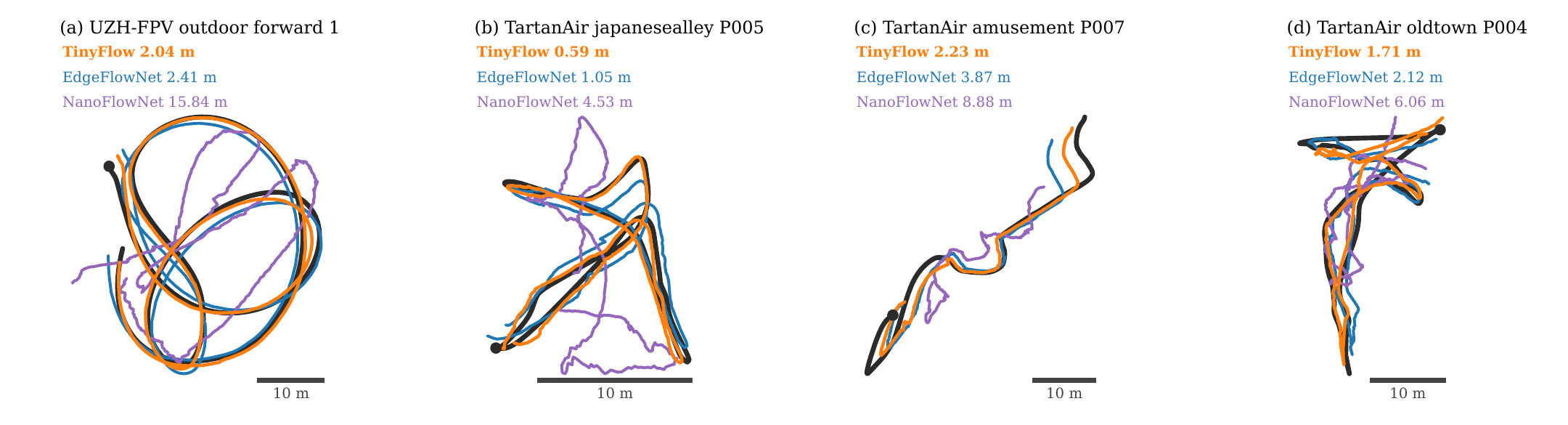}
\caption{\textbf{Flow-only visual odometry trajectories at INT8.} Estimated paths of TinyFlow (orange), EdgeFlowNet (blue) and NanoFlowNet (purple) against the ground truth (black) for one UZH-FPV flight (a) and three TartanAir trajectories (b)--(d), with each model's ATE from Table~\ref{tab:vo}. The black dot marks the start of the trajectory. TinyFlow stays on the ground-truth path through loops and sharp turns, EdgeFlowNet drifts on the return legs, and NanoFlowNet leaves the path.}
\label{fig:vo_fig_qual}
\vspace{-10pt}
\end{figure*}

\begin{table}[t!]
\centering
\caption{Flow-only monocular visual odometry with the INT8 models on eight
TartanAir trajectories and three UZH-FPV flights. The estimation method is
described in Sec.~\ref{sec:vo}. ATE is the position RMSE in metres after
SE(3) alignment. Rot.\ is the median per-frame rotation error and Dir.\ the
mean per-frame translation direction error, both in degrees. GT is the
rendered ground-truth flow through the same estimator. Means are unweighted
over sequences. Bold marks the best deployable value in each column.
TartanAir: factory = abandonedfactory P001, amuse.\ = amusement P007,
gascola = gascola P001, alley = japanesealley P005, neighb.\ = neighborhood
P002, oldtown = oldtown P004, forest = seasonsforest P003, soulcity =
soulcity P003. UZH-FPV: in\_45\_2 = indoor\_45\_2, in\_fwd\_3 =
indoor\_forward\_3, out\_fwd\_1 = outdoor\_forward\_1.}
\label{tab:vo}
\footnotesize
\setlength{\tabcolsep}{1.6pt}
\renewcommand{\arraystretch}{1.15}
\setlength{\aboverulesep}{0.4ex}\setlength{\belowrulesep}{0.65ex}
\begin{tabular*}{\columnwidth}{@{\extracolsep{\fill}}l rrr rrr rrr r@{}}
\toprule
 & \multicolumn{3}{c}{\textbf{TinyFlow (Ours)}} & \multicolumn{3}{c}{EdgeFlowNet} & \multicolumn{3}{c}{NanoFlowNet} & GT \\
\cmidrule(lr){2-4}\cmidrule(lr){5-7}\cmidrule(lr){8-10}\cmidrule(l){11-11}
Seq. & ATE$\downarrow$ & Rot.$\downarrow$ & Dir.$\downarrow$ & ATE$\downarrow$ & Rot.$\downarrow$ & Dir.$\downarrow$ & ATE$\downarrow$ & Rot.$\downarrow$ & Dir.$\downarrow$ & ATE \\
\midrule
\multicolumn{11}{c}{\emph{TartanAir, synthetic}}\\
\cmidrule(lr){1-11}
factory & 5.20 & \textbf{0.31} & \textbf{28.4} & \textbf{3.49} & 0.37 & 29.6 & 9.36 & 0.99 & 56.8 & 0.80 \\
amuse. & \textbf{2.23} & \textbf{0.21} & \textbf{25.7} & 3.87 & \textbf{0.21} & 28.8 & 8.88 & 0.56 & 47.2 & 0.62 \\
gascola & \textbf{1.72} & \textbf{0.21} & \textbf{31.9} & 1.88 & \textbf{0.21} & 37.3 & 3.86 & 0.46 & 54.5 & 0.09 \\
alley & \textbf{0.59} & \textbf{0.27} & \textbf{14.5} & 1.05 & 0.29 & 19.2 & 4.53 & 0.98 & 44.3 & 0.07 \\
neighb. & \textbf{4.88} & 0.30 & 34.2 & 5.37 & \textbf{0.28} & \textbf{33.0} & 11.72 & 0.95 & 62.7 & 0.27 \\
oldtown & \textbf{1.71} & \textbf{0.26} & \textbf{22.5} & 2.12 & 0.30 & 25.8 & 6.06 & 0.78 & 48.9 & 0.13 \\
forest & \textbf{0.72} & 0.26 & 14.5 & 0.73 & \textbf{0.24} & \textbf{13.0} & 3.13 & 0.66 & 33.8 & 0.04 \\
soulcity & 2.60 & 0.44 & \textbf{19.8} & \textbf{1.99} & \textbf{0.40} & 21.9 & 5.68 & 1.10 & 44.6 & 0.04 \\
\cmidrule{1-11}
\emph{Mean} & \textbf{2.46} & \textbf{0.28} & \textbf{23.9} & 2.56 & 0.29 & 26.1 & 6.65 & 0.81 & 49.1 & 0.26 \\
\midrule
\multicolumn{11}{c}{\emph{UZH-FPV, real drone flights}}\\
\cmidrule(lr){1-11}
in\_45\_2 & 9.37 & \textbf{0.60} & \textbf{22.9} & \textbf{8.21} & \textbf{0.60} & 25.0 & 10.86 & 3.00 & 42.3 & -- \\
in\_fwd\_3 & 4.53 & \textbf{0.33} & 9.2 & \textbf{2.14} & 0.34 & \textbf{6.7} & 9.24 & 0.78 & 32.3 & -- \\
out\_fwd\_1 & \textbf{2.04} & \textbf{0.16} & \textbf{4.5} & 2.41 & 0.17 & 4.8 & 15.84 & 0.44 & 25.9 & -- \\
\cmidrule{1-11}
\emph{Mean} & 5.32 & \textbf{0.37} & \textbf{12.2} & \textbf{4.25} & \textbf{0.37} & \textbf{12.2} & 11.98 & 1.41 & 33.5 & -- \\
\bottomrule
\end{tabular*}
\vspace{-15pt}
\end{table}
\raggedbottom

We deploy these models on a Raspberry Pi 5 with a Hailo-8 AI accelerator. On this setup, TinyFlow (1.26\,M parameters) runs at 10.5\,FPS with 5.76 EPE on a 100-pair Sintel Clean sample, measured on the same chip and frames as EdgeFlowNet (2.75\,M, 64\,FPS, 6.38 EPE) and NanoFlowNet (0.17\,M, 190\,FPS, 7.50 EPE). Iterative refinement therefore improves accuracy at the cost of latency while the parameter count remains below that of EdgeFlowNet.

\subsection{Visual Odometry Evaluation}\label{sec:vo}
Dense flow on a small robot is rarely the end product, so we test whether the ranking of Table~\ref{tab:flow_accuracy} survives a downstream task. Table~\ref{tab:vo} reports flow-only monocular visual odometry (VO) on eight synthetic TartanAir trajectories~\cite{tartanair} and three real UZH-FPV drone flights~\cite{uzhfpv}. Every deployable model runs at INT8 and feeds the same five-point RANSAC pose estimator, no depth is used, and the translation scale is taken from the ground truth, so the metrics measure only how well the flow constrains the camera rotation and the direction of travel. On TartanAir, TinyFlow has the lowest ATE on six of the eight sequences and the lowest mean on all three metrics, 2.46\,m, 0.28$^\circ$ and 23.9$^\circ$ against 2.56\,m, 0.29$^\circ$ and 26.1$^\circ$ for EdgeFlowNet, with the largest margins on japanesealley\_P005 and amusement\_P007, where its ATE is 44\% and 42\% lower. NanoFlowNet drifts 2.7$\times$ as far as TinyFlow and doubles its direction error, so the structured noise seen in Fig.~\ref{fig:fig_qual} turns directly into pose error. The rendered ground-truth flow through the same estimator gives 0.26\,m mean ATE, an order of magnitude below every learned model, so the remaining error lies in the flow rather than in the estimator. On the real flights the result is mixed. TinyFlow is best on all three metrics on outdoor\_forward\_1 (2.04\,m against 2.41\,m) and the two models tie in mean rotation and direction error (0.37$^\circ$, 12.2$^\circ$), but EdgeFlowNet drifts less on the two indoor flights and has the lower mean ATE, 4.25\,m against 5.32\,m. Over all eleven sequences TinyFlow has the lowest ATE on seven, EdgeFlowNet on four and NanoFlowNet on none.

Fig.~\ref{fig:vo_fig_qual} overlays four of the estimated paths on the ground truth. TinyFlow follows the repeated loops of the outdoor flight (a) and the sharp turns of japanesealley (b), where EdgeFlowNet is offset on the return legs; on amusement (c) EdgeFlowNet veers off near the end of the route, and on oldtown (d) both models follow the route with TinyFlow closer at the turns. NanoFlowNet wanders off the path in every panel. The VO results thus confirm the EPE ranking: among the deployable models, TinyFlow produces the flow most consistent with the camera motion at the smallest parameter count.

\subsection{Robustness to Illumination Change}
\label{sec:robust}
\begin{figure}[t!]
\centering
\includegraphics[width=\columnwidth]{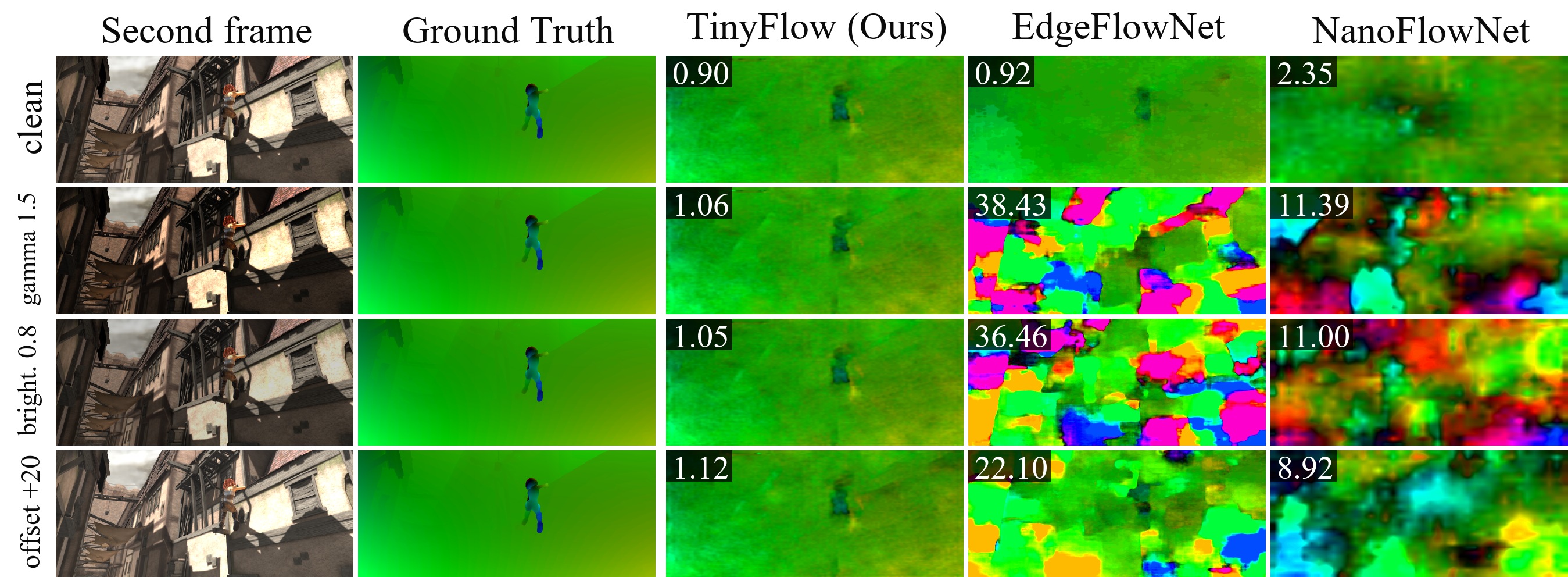}
\caption{One Sintel Clean pair, unchanged and with gamma 1.5, brightness
scaling by 0.8 and an offset of $+$20 applied to the second frame. The
first column shows the second frame each model receives. All flow models run INT8 variants.}
\label{fig:illum}
\vspace{-15pt}
\end{figure}

Benchmarks render both frames under identical lighting, whereas
auto-exposure on a moving robot changes the brightness of the second frame
relative to the first. We apply gamma, brightness scaling and an additive
offset to the second frame of a fixed 209-pair Sintel Clean subset and
leave the ground truth untouched (Table~\ref{tab:illum}). TinyFlow stays
within 5.92 to 8.99 EPE across all eight conditions. EdgeFlowNet rises to
27 to 61 and NanoFlowNet to 18 to 90. Fig.~\ref{fig:illum} shows the failure on one frame.
TinyFlow moves from 0.90 to at most 1.12 EPE under the three changes, while
EdgeFlowNet predicts flow up to four times the true magnitude. We attribute this to the
$\tanh$-bounded update heads, which limit the flow TinyFlow can emit
regardless of how far the features shift. With these experiments, we demonstrate that TinyFlow is robust to such illumination changes.

\begin{table}[t!]
\centering
\caption{EPE of the INT8 models under illumination change applied to the
second frame only, on a fixed 209-pair Sintel Clean subset. We test various gamma, brightness scaling and image offset levels. NanoFlowNet runs at its native
$112{\times}160$ as in Table \ref{tab:flow_accuracy}.}

\label{tab:illum}
\scriptsize
\setlength{\tabcolsep}{2.2pt}
\setlength{\aboverulesep}{0.25ex}\setlength{\belowrulesep}{0.35ex}
\begin{tabular}{@{}l c cccc cc cc@{}}
\toprule
 & & \multicolumn{4}{c}{Gamma} & \multicolumn{2}{c}{Brightness} & \multicolumn{2}{c}{Offset} \\
\cmidrule(lr){3-6}\cmidrule(lr){7-8}\cmidrule(l){9-10}
Model & None & 0.70 & 0.85 & 1.20 & 1.50 & 0.80 & 1.25 & $+$20 & $-$20 \\
\midrule
EdgeFlowNet & 6.02 & 60.92 & 29.00 & 41.94 & 60.05 & 40.38 & 26.99 & 52.05 & 50.55 \\
NanoFlowNet & 7.73 & 90.12 & 17.78 & 23.17 & 61.68 & 21.16 & 20.24 & 75.15 & 41.38 \\
\textbf{TinyFlow (Ours)} & \textbf{5.65} & \textbf{6.78} & \textbf{5.92} & \textbf{6.08} & \textbf{6.67} & \textbf{6.29} & \textbf{8.99} & \textbf{6.29} & \textbf{6.57} \\
\bottomrule
\end{tabular}
\end{table}

\vspace{-5pt}
\section{Conclusion}
TinyFlow is an edge-deployable 1.26M-parameter optical flow network that retains iterative refinement and removes warping. Its fixed-window correlation carries the data dependence of warping in channel values rather than in memory addresses. We demonstrate that TinyFlow outperforms EdgeFlowNet on Sintel Clean, Sintel Final, and FlyingChairs in both precisions with 54\% fewer parameters, and leads NanoFlowNet by a wide margin at every motion magnitude.
The ranking carries over to flow-only visual odometry, where TinyFlow has the lowest ATE on seven of the eleven TartanAir and UZH-FPV sequences.

\bibliographystyle{IEEEtran}
\bibliography{references}

@article{chao2014survey,
  title   = {A Survey of Optical Flow Techniques for Robotics Navigation Applications},
  author  = {Chao, Haiyang and Gu, Yu and Napolitano, Marcello},
  journal = {Journal of Intelligent \& Robotic Systems},
  volume  = {73},
  number  = {1},
  pages   = {361--372},
  year    = {2014}
}

@article{pandey2021leveraging,
  title   = {Leveraging Deep Learning for Visual Odometry Using Optical Flow},
  author  = {Pandey, Tejas and Pena, Dexmont and Byrne, Jonathan and Moloney, David},
  journal = {Sensors},
  volume  = {21},
  number  = {4},
  pages   = {1313},
  year    = {2021}
}

@inproceedings{schonberger2016structure,
  title     = {Structure-from-Motion Revisited},
  author    = {Sch{\"o}nberger, Johannes L. and Frahm, Jan-Michael},
  booktitle = {IEEE Conference on Computer Vision and Pattern Recognition (CVPR)},
  pages     = {4104--4113},
  year      = {2016}
}

@article{karpuzov2024object,
  title   = {Object Tracking Based on Optical Flow Reconstruction of Motion-Group Parameters},
  author  = {Karpuzov, Simeon and Petkov, George and Ilieva, Sylvia and Petkov, Alexander and Kalitzin, Stiliyan},
  journal = {Information},
  volume  = {15},
  number  = {6},
  pages   = {296},
  year    = {2024}
}

@phdthesis{luo2022motion,
  title  = {Motion Learning for Dynamic Scene Understanding},
  author = {Luo, Chenxu},
  school = {Johns Hopkins University},
  year   = {2022}
}

@inproceedings{bouwmeester2023nanoflownet,
  title     = {{NanoFlowNet}: Real-Time Dense Optical Flow on a Nano Quadcopter},
  author    = {Bouwmeester, Rik J. and Paredes-Vall{\'e}s, Federico and de Croon, Guido C. H. E.},
  booktitle = {IEEE International Conference on Robotics and Automation (ICRA)},
  pages     = {1996--2003},
  year      = {2023}
}

@article{singh2024minimal,
  title   = {Minimal Perception: Enabling Autonomy in Resource-Constrained Robots},
  author  = {Singh, Chahat Deep and He, Botao and Ferm{\"u}ller, Cornelia and Metzler, Christopher and Aloimonos, Yiannis},
  journal = {Frontiers in Robotics and AI},
  volume  = {11},
  pages   = {1431826},
  year    = {2024}
}

@article{edgeflownet2024,
  title   = {{EdgeFlowNet}: 100{FPS}@1{W} Dense Optical Flow for Tiny Mobile Robots},
  author  = {{Pinnama Raju}, Sai Ramana Kiran and Singh, Rishabh and Velmurugan, Manoj and Sanket, Nitin J.},
  journal = {IEEE Robotics and Automation Letters},
  volume  = {10},
  number  = {1},
  pages   = {128--135},
  year    = {2025}
}

@article{krispin2024real,
  title   = {Real-Time {3D} Object Detection Using {InnovizOne} {LiDAR} and Low-Power {Hailo-8} {AI} Accelerator},
  author  = {Krispin-Avraham, Itay and Orfaig, Roy and Bobrovsky, Ben-Zion},
  journal = {arXiv preprint arXiv:2412.05594},
  year    = {2024}
}

@inproceedings{dosovitskiy2015flownet,
  title     = {{FlowNet}: Learning Optical Flow with Convolutional Networks},
  author    = {Dosovitskiy, Alexey and Fischer, Philipp and Ilg, Eddy and H{\"a}usser, Philipp and Hazirbas, Caner and Golkov, Vladimir and van der Smagt, Patrick and Cremers, Daniel and Brox, Thomas},
  booktitle = {IEEE International Conference on Computer Vision (ICCV)},
  pages     = {2758--2766},
  year      = {2015}
}

@inproceedings{ilg2017flownet,
  title     = {{FlowNet 2.0}: Evolution of Optical Flow Estimation with Deep Networks},
  author    = {Ilg, Eddy and Mayer, Nikolaus and Saikia, Tonmoy and Keuper, Margret and Dosovitskiy, Alexey and Brox, Thomas},
  booktitle = {IEEE Conference on Computer Vision and Pattern Recognition (CVPR)},
  pages     = {2462--2470},
  year      = {2017}
}

@inproceedings{sun2018pwc,
  title     = {{PWC-Net}: {CNNs} for Optical Flow Using Pyramid, Warping, and Cost Volume},
  author    = {Sun, Deqing and Yang, Xiaodong and Liu, Ming-Yu and Kautz, Jan},
  booktitle = {IEEE/CVF Conference on Computer Vision and Pattern Recognition (CVPR)},
  pages     = {8934--8943},
  year      = {2018}
}

@inproceedings{teed2020raft,
  title     = {{RAFT}: Recurrent All-Pairs Field Transforms for Optical Flow},
  author    = {Teed, Zachary and Deng, Jia},
  booktitle = {European Conference on Computer Vision (ECCV)},
  pages     = {402--419},
  year      = {2020}
}

@inproceedings{hui2020liteflownet3,
  title     = {{LiteFlowNet3}: Resolving Correspondence Ambiguity for More Accurate Optical Flow Estimation},
  author    = {Hui, Tak-Wai and Tang, Chen Change},
  booktitle = {European Conference on Computer Vision (ECCV)},
  pages     = {169--184},
  year      = {2020}
}

@inproceedings{kong2021fastflownet,
  title     = {{FastFlowNet}: A Lightweight Network for Fast Optical Flow Estimation},
  author    = {Kong, Lingtong and Shen, Chunhua and Yang, Jie},
  booktitle = {IEEE International Conference on Robotics and Automation (ICRA)},
  pages     = {10310--10316},
  year      = {2021}
}

@inproceedings{jiang2023dcvnet,
  title     = {{DCVNet}: Dilated Cost Volume Networks for Fast Optical Flow},
  author    = {Jiang, Huaizu and Learned-Miller, Erik},
  booktitle = {IEEE/CVF Winter Conference on Applications of Computer Vision (WACV)},
  pages     = {5139--5146},
  year      = {2023}
}

@inproceedings{zhang2025neuflow,
  title     = {{NeuFlow-V2}: Push High-Efficiency Optical Flow to the Limit},
  author    = {Zhang, Zhiyong and Gupta, Aniket and Jiang, Huaizu and Singh, Hanumant},
  booktitle = {IEEE/RSJ International Conference on Intelligent Robots and Systems (IROS)},
  pages     = {2479--2485},
  year      = {2025}
}

@inproceedings{jacob2018quantization,
  title     = {Quantization and Training of Neural Networks for Efficient Integer-Arithmetic-Only Inference},
  author    = {Jacob, Benoit and Kligys, Skirmantas and Chen, Bo and Zhu, Menglong and Tang, Matthew and Howard, Andrew and Adam, Hartwig and Kalenichenko, Dmitry},
  booktitle = {IEEE Conference on Computer Vision and Pattern Recognition (CVPR)},
  pages     = {2704--2713},
  year      = {2018}
}

@article{hinton2015distilling,
  title   = {Distilling the Knowledge in a Neural Network},
  author  = {Hinton, Geoffrey and Vinyals, Oriol and Dean, Jeff},
  journal = {arXiv preprint arXiv:1503.02531},
  year    = {2015}
}

@inproceedings{wortsman2022model,
  title     = {Model Soups: Averaging Weights of Multiple Fine-Tuned Models Improves Accuracy without Increasing Inference Time},
  author    = {Wortsman, Mitchell and Ilharco, Gabriel and Gadre, Samir Yitzhak and Roelofs, Rebecca and Gontijo-Lopes, Raphael and Morcos, Ari S. and Namkoong, Hongseok and Farhadi, Ali and Carmon, Yair and Kornblith, Simon and Schmidt, Ludwig},
  booktitle = {International Conference on Machine Learning (ICML)},
  pages     = {23965--23998},
  year      = {2022}
}

@inproceedings{reuther2022survey,
  title     = {{AI} and {ML} Accelerator Survey and Trends},
  author    = {Reuther, Albert and Michaleas, Peter and Jones, Michael and Gadepally, Vijay and Samsi, Siddharth and Kepner, Jeremy},
  booktitle = {IEEE High Performance Extreme Computing Conference (HPEC)},
  year      = {2022}
}

@inproceedings{mayer2016things3d,
  title     = {A Large Dataset to Train Convolutional Networks for Disparity, Optical Flow, and Scene Flow Estimation},
  author    = {Mayer, Nikolaus and Ilg, Eddy and H{\"a}usser, Philipp and Fischer, Philipp and Cremers, Daniel and Dosovitskiy, Alexey and Brox, Thomas},
  booktitle = {IEEE Conference on Computer Vision and Pattern Recognition (CVPR)},
  pages     = {4040--4048},
  year      = {2016}
}

@inproceedings{tartanair,
  title     = {{TartanAir}: A Dataset to Push the Limits of Visual {SLAM}},
  author    = {Wang, Wenshan and Zhu, Delong and Wang, Xiangwei and Hu, Yaoyu and Qiu, Yuheng and Wang, Chen and Hu, Yafei and Kapoor, Ashish and Scherer, Sebastian},
  booktitle = {IEEE/RSJ International Conference on Intelligent Robots and Systems (IROS)},
  pages     = {4909--4916},
  year      = {2020}
}

@inproceedings{uzhfpv,
  title     = {Are We Ready for Autonomous Drone Racing? {The} {UZH-FPV} Drone Racing Dataset},
  author    = {Delmerico, Jeffrey and Cieslewski, Titus and Rebecq, Henri and Faessler, Matthias and Scaramuzza, Davide},
  booktitle = {IEEE International Conference on Robotics and Automation (ICRA)},
  pages     = {6713--6719},
  year      = {2019}
}
\end{document}